\pdfoutput=1

\documentclass[11pt]{article}

\usepackage[preprint]{acl}

\usepackage{times}
\usepackage{latexsym}
\usepackage{multirow}
\usepackage[T1]{fontenc}
\usepackage{subfigure}
\usepackage[utf8]{inputenc}
\usepackage{pifont}
\usepackage{microtype}
\usepackage{booktabs}
\usepackage{makecell}
\usepackage{graphicx}
\usepackage{amsmath,amssymb,amsfonts}
\usepackage{xspace}
\usepackage{inconsolata}
\usepackage[listings]{tcolorbox}
\usepackage{float}

\author{Haitao Mao$^{1}$, Guangliang Liu$^{1}$, Yao Ma$^{2}$, Rongrong Wang$^{1}$, Kristen Johnson$^{1}$, Jiliang Tang$^{1}$ \\
$^{1}$Michigan State University $^{2}$Rensselaer Polytechnic Institute \\
\texttt{\{haitaoma,liuguan5,wangron6,kristenj,tangjili\}@msu.edu} \\
\texttt{may13@rpi.edu}
  }

\title{A Survey to Recent Progress Towards Understanding In-Context Learning}

\begin{document}

\maketitle

\begin{abstract}
In-Context Learning (ICL) empowers Large Language Models (LLMs) with the ability to learn from a few examples provided in the prompt, enabling downstream generalization without the requirement for gradient updates. Despite encouragingly empirical success, the underlying mechanism of ICL remains unclear. Existing research remains ambiguous with various viewpoints, utilizing intuition-driven and ad-hoc technical solutions to interpret ICL. In this paper, we leverage a data generation perspective to reinterpret recent efforts from a systematic angle, demonstrating the potential broader usage of these popular technical solutions. For a conceptual definition, we rigorously adopt the terms of \textit{skill recognition} and \textit{skill learning}. Skill recognition selects one learned data generation function previously seen during pre-training while skill learning can learn new data generation functions from in-context data. Furthermore, we provide insights into the strengths and weaknesses of both abilities, emphasizing their commonalities through the perspective of data generation. This analysis suggests potential directions for future research.
\end{abstract}

\section{Introduction}

LLMs have revolutionized Natural Language Processing~(NLP)~\cite{achiam2023gpt} and other relevant areas such as multi-modal tasks over vision and language~\cite{liu2023llava}, accelerating numerous challenging research directions, e.g., AI agent~\cite{durante2024agent}, reasoning~\cite{wei2022chain}, and story telling~\cite{xie2023next}. 
These amazing applications display LLMs’ emerging capabilities, which can be formally defined as new abilities that are not present in small models but arise in larger ones~\cite{zhao2023survey}. 
Among them, the emerging ICL ability serves as an important foundation of other capabilities. 
Notably, small models also have the capability to perform ICL, but the level of capability is different from that of larger models, wherein people can easily observe more in-depth displays of understanding for the given context of inputs, e.g., identify long-term dependency and abstract concept comprehension. 
For instance, \citet{ganguli2023capacity} demonstrates that only LLMs over 22B parameters can understand the moral concepts, being able to generate unbiased answers. 

\begin{figure}
    \centering
    \includegraphics[width=0.48\textwidth]{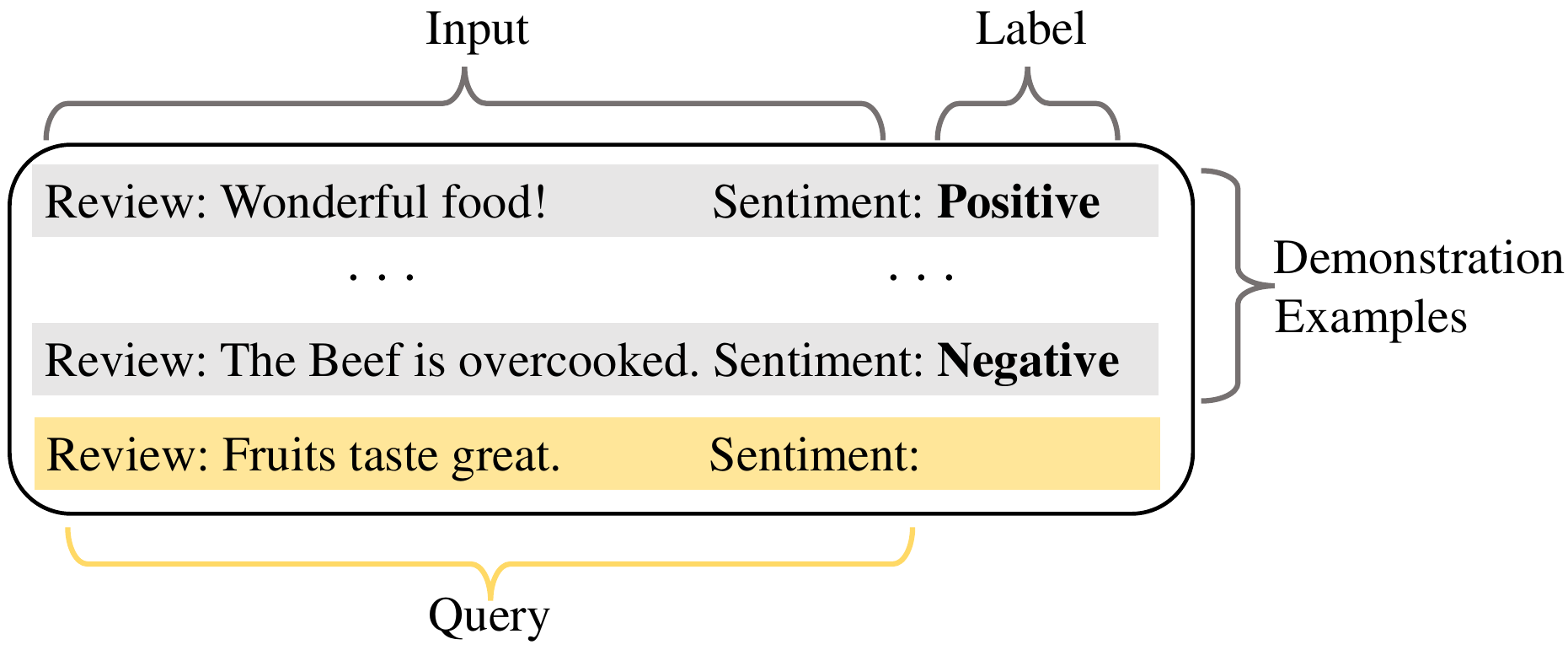}
    \vskip -0.7em
    \caption{\small Illustration of ICL for Sentiment Analysis. The upper instances (with background color gray) are the labeled in-context demonstrations, while the last line is the query for which LLMs infer the sentiment label. \label{fig:ICL-demonstration}}
    \vskip -1.6em
\end{figure}

ICL, a fundamental and emerging capability serving as the pre-requisite for many complicated abilities, is the process of leveraging a few selected labeled demonstrations with the format \textit{(input, label)}\footnote{In this paper, we focus on classification tasks, as they are widely used in theoretical studies of ICL due to their well-defined mathematical tools and clear evaluation metrics.}, before the query input, for making predictions in a few-/one-shot manner. 
An example of ICL is illustrated in Figure~\ref{fig:ICL-demonstration}. 

\begin{table*}[htb]
    \caption{\small A summarization table of representative works. SR and SL stand for skill recognition and skill learning, respectively. Function approximation revolves on how effectively ICL can fit different generalize functions. The Internal Mechanism describes how LLMs learn through various gradient descent algorithms. More details on empirical simplification and theoritical assumptions can be found in Appendix~\ref{app:simplification}.
    \label{tab:summarization}}
    \resizebox{\linewidth}{!}{
    \begin{tabular}{lllll}
    \hline
        Literature & Ability & Analysis View & Date Generation Function & Characteristics \\ \hline
        \multirow{2}{*}{\citet{xie2021explanation,zhang2023and}} & \multirow{2}{*}{SR} & Theoretical & \multirow{2}{*}{HMM} & \multirow{2}{*}{Internal Mechanism} \\ 
        &  &\& Empirical & &  \\ 
        \citet{wang2023large} & SR & Empirical & LDA & Generalization \\ 
        \citet{zhao2023context} & SR & Theoretical & Hopfield Network & Internal Mechanism \\ 
        \citet{raventos2023pretraining} & SL & Theoretical & linear regression & Generalization \\ 
        \citet{wu2023many} & SL & Empirical & linear regression & Generalization \\ 
        \citet{garg2022can} & SL & Empirical & linear regression, decision tree, NN & Function Approximation \\ 
        \citet{bai2023transformers,fu2023can}  & SL & Theoretical & linear regression, decision tree, NN & Generalization \\ 
        \citet{yadlowsky2023pretraining, ahuja2023context} & SL & Empirical & linear regression, polynomial regression & Generalization \\ 
    \citet{von2023transformers,zhang2023can}  &  \multirow{2}{*}{SL} & \multirow{2}{*}{Theoretical} & \multirow{2}{*}{linear regression} & \multirow{2}{*}{Internal Mechanism} \\ 
        \cite{mahankali2023one, ahn2023transformers} & \\ 
        
        \citet{akyurek2022learning} & SL & Theoretical & linear regression & Internal Mechanism \\ 
    \citet{li2023shifting, ren2023context} & \multirow{2}{*}{SL} & \multirow{2}{*}{Theoretical} & \multirow{2}{*}{non-linear regression} & \multirow{2}{*}{Internal Mechanism} \\ 
        \citet{cheng2023transformers,guo2023transformers} & \\
        \citet{hahn2023theory} & SR\&SL & Theoretical & context-free grammar & Generalization \\ \hline
    \end{tabular}
    }
    \vspace{-1em}
\end{table*}

Despite the empirical success of various ICL prompting strategies for downstream applications~\cite{mavromatis2023examples, ye2022complementary}, the mechanism of ICL remains unclear, leading to unexplainable observations, e.g., sensitivity to the sample order~\cite{lu2021fantastically}, or being robust to human-crafted yet irrational input-label mapping.
Increasing attention has been paid to understand ICL from various perspectives. 
However, this area is still growing, with many open research questions are actively being explored. 
Due to the complexity of LLMs, most existing works only take one individual factor into account, e.g., the pre-training data distribution~\cite{chan2022data}, model scale~\cite{wei2023larger}, or difficulty level of the in-context task~\cite{raventos2023pretraining}. 
Moreover, existing works focusing the same factor may adopt different experimental settings~\cite{yoo2022ground, min2022rethinking}, leading to potentially conflicting conclusions. 
Typically, \citet{pan2023context} categorizes ICL into two abilities: task recognition and task learning.

In this paper, we propose the data generation perspective as a principled angle to comprehend existing studies towards understanding ICL. 
Following this perspective, the pretraining stage can be interpreted as learning the data generation function classes underlying pretraining corpus, where the masked language modeling objective~\cite{devlin2019bert} and the next token prediction objective~\cite{radford2018improving} are both objectives that allow us learn the data generation functions.
Similarly, the ICL stage can be considered as a label generation process given the query inputs. 
Therefore, adopting this data generation perspective enables a unified framework through which we can cohesively analyze both pretraining and ICL stages, offering a holistic approach to understanding the foundations of LLMs.

Guided by the data generation perspective, we introduce a more principled and rigorous understanding framework on \textit{skill learning} and \textit{skill recognition}, distinguished by whether LLMs can learn a new data generation function in context. 
The skill learning ability is to learn a new data generation function in context, which is unseen in the pretraining stage. 
The skill recognition ability selects one learned data generation function previously seen during pre-training. 
To analyze the mechanism of abilities, the function learning statistical framework~\cite{garg2022can} and the Bayesian inference statistical framework~\cite{xie2021explanation} are representative works for skill learning and skill recognition ability, respectively.

\textbf{Organization: } Section~\ref{sec:related} introduces previous studies of ICL and Section~\ref{sec:term} presents the terminology. 
Key contributions lie in Section~\ref{sec:skill-recognition} and~\ref{sec:skill-learning}, which systematically review the skill recognition with the Bayesian inference framework and the skill learning with the function learning framework, respectively. 
We outline the challenges and potential directions in Section~\ref{sec:insight}, aiming to offer a valuable guide for newcomers to the field while also illuminating pathways for future research.

\section{Related works\label{sec:related}}

\textbf{Comparison with existing relevant literature}. 
The key difference between our work and existing ones lies in its more dedicating scope on the mechanism of ICL and advocating a principled data generation perspective, instead of a broad, application-oriented perspective in~\citet{dong2022survey, zhao2023survey, wei2022emergent}. Our work provides a comprehensive literature review and clear categorization as shown in Table~\ref{tab:summarization}. 
Moreover, we propose a new holistic data generation perspective which can be utilized for the tokenizer, which clarifies the connections and distinctions between different data statistical frameworks. 

\textbf{Distinguish skill learning from skill recognition}. 
The skill can be regarded as a data generation function, referring to the underlying hypothesis on the textual data generation.
To determine whether the utilized skill is from the pre-training function class or is a new function, an empirical method is to validate whether LLMs can fit a set of data generated with a ground-truth function which is outside the pre-training function class.

\textbf{Distinguish skill recognition/learning from task recognition/learning}~\cite{pan2023context}. We distinguish our proposed skill recognition/learning from a data generation perspective with previous task recognition/learning proposed in~\cite{pan2023context}. 
Task recognition/learning is a narrower aspect of our skill recognition/learning as they majorly focus on the empirical performance variation under the label permutation on in-context data. Task learning is recognized as performance degradation, indicating ICL learns the permuted in-context data. In contrast, the task recognition corresponds to the unchanged performance, indicating ICL only relies on pre-training knowledge. 
The key advantages of our proposed skill recognition/learning definition are shown as follows: 
(1) Thanks to the mathematical description with a data generation function, skill learning/recognition enables both theoretical analysis and empirical evidence, instead of only focusing on the empirical one. 
(2) Task recognition/learning can only emphasize the performance of a classification task in complicated real-world applications. Instead, skill learning/recognition can utilize different existing data generation functions in the NLP domain, e.g., HMM, and LDA, rather than merely input-label mapping for classification. Moreover, the data generation enables to conduct synthetic analyses in a systematic and controllable setting.

\section{Terminology\label{sec:term}} 

The prompt sequence of In-Context Learning consists of two parts: 
(1) The demonstration is illustrated as an \textit{(input, label)} pair, denoted as $(x_i, y_i)$; These demonstrations provide the basic description of the intended task.
(2) The query is the test input after a few demonstrations. ICL aims to provide the correct prediction for the query based on the in-context demonstrations and the prior knowledge of a pre-trained LLM. 
The \textit{data generation function} in this paper refers to the underlying hypothesis on language data generation. 
It serves as the data assumption in the theoretical understanding and the simulation data generator for the synthetic experimental analysis. 
Each data generation function obtained by the LLM can be recognized as a skill.

\section{Skill Recognition\label{sec:skill-recognition}}

Skill recognition ability is the ability of an LLM to select the most proper data generation function from the function class obtained during pre-training.
And this selection process is driven by the in-context demonstrations.
A Bayesian inference framework~\cite{xie2021explanation} is introduced to explain the skill recognition. 
The ICL inference can be instantiated as a Bayesian inference process as follows: 
 \begin{equation*}
    \begin{split}
        &p(\text{y}|\text{prompt})=\\ & \int_{\text{concept}} \!\!p(\text{y}|\text{concept},\!\text{prompt}) p(\text{concept}|\text{prompt})d(\text{concept})
    \end{split} 
    \label{eq:bayesian}
\end{equation*}
    
where $p(\text{y}|\text{prompt})$ is the conditional probability of the output generation $\text{y}$ given the prompt. 
It can be marginalized with pre-training concepts and \textit{each concept corresponds to a pre-training data generation function}. 
$p(\text{concept}|\text{prompt})$ is the probability of locating the latent concept aligned with in-context demonstrations.
After locating the aligned concept, $p(\text{y}|\text{concept},\text{prompt})$ utilizes the selected data generation function for the output generation.

This approach to modeling latent concepts is widely used in the field of NLP, as language data is inherently compositional, involving underlying concepts—such as sentiment, topics, and syntactic structures—that are not explicitly observable in the raw text~\cite{chung2015recurrent,zhou2020towards}.
Latent variable models can specify prior knowledge and structural dependencies for language data which enjoys the characteristics of high compositionality. 
Deep latent variable models are popularly utilized to improve various tasks such as alignment in statistical machine translation, topic modeling, and text generation~\cite{kim2018tutorial,fang2019implicit,wang2023large}.

Though there are various definitions of latent concepts, any latent information that can help ICL can be considered as a good choice for the \textit{concept} in the Bayesian inference process above.
We summarize the existing concept definitions as follows: 
(1) \citet{xie2021explanation} defines the concept as the transition matrix $\mathbf{\theta}$ of a Hidden Markov Model~(HMM)~\cite{baum1966statistical}, which assumes to be the underlying distribution of the real-world language data. 
The concept helps to state a transition distribution over observed tokens. 
A concrete example of the concept is the transition between name (Albert Einstein) $\to$ nationality (German) $\to$ occupation (physicist) in wiki bios.
(2) \citet{wang2023large} simplifies the transition between tokens, modeled by HMM, with LDA topic models where each topic corresponds to one latent concept~\cite{blei2003latent}. 
(3) Despite the above mathematical interpretations, \citet{todd2023function} and \citet{liu2023context} empirically establish the connection between the latent concept and the downstream task, e.g., supervised classification and question-answering, where the particular latent representation in the LLM can capture essential information about the task.

The Bayesian inference framework is firstly proposed by~\citet{xie2021explanation}, interpreting how obtained pre-training data functions are activated by in-context demonstrations. 
Key challenges in this framework are:
(1) In the pre-training stage, how the model obtains the latent concepts from the pre-training corpus; and  
(2) In the ICL inference stage, how in-context demonstrations can locate the most relevant concept to generate the desired output. 

The pre-training stage aims to obtain various concepts from the large pre-training corpora if each pre-training document is generated from an individual HMM model. 
In such cases, the next token prediction objective can converge if and only if the LLM can successfully generate the correct next token matching the HMM transitions. The transitions are dominated by the underlying concept~\cite{xie2021explanation}.  
Different documents can be generated from various concepts sampled from the concept set denoted as $\Theta$.

The ICL inference stage conducts an implicit Bayesian inference to locate an appropriate concept $\theta^* \in \Theta$ which shows the  optimal likelihood to generate the given in-context demonstrations. The format of the prompt is shown below:
\begin{equation}
\begin{split}
     &\left [ S_n, x_{\text{test}} \right ]\\
     &= \left[x_{1}, y_{1}, o^{\text{del}}, \ldots, x_{n}, y_{n}, o^{\text{del}}, x_{\text{test}}\right] \sim p_{\text{prompt}}
\end{split}
\end{equation}
where $p_{\text{prompt}}$ is a data generation process implemented with HMM parameterized by $\theta^*$. 
$x_i$, $y_i$ and $o^{\text{del}}$ are the input, label, and delimiter, respectively. 
The difficulty in locating $\theta^*$ is due to low probability for all the pre-training concepts to generate the in-context demonstrations. 
The key reason is that token transition patterns of the in-context demonstrations are of three types: (1) the input to the label $x_i\to y_i$, (2) the label to the delimiter, and (3) the delimiter to the input. 
The latter two patterns hardly appear in the pre-training data due to different delimiter usages. 

To address the above issue of low probability, \citet{xie2021explanation} proposes some assumptions. 
One example is the located concept $\theta^*$ enjoys a higher probability transiting to delimiters than that of other concepts. 
Equipped with those assumptions, we are able to locate the aligned pre-training concept to implement Bayesian inference. 
The model can locate the corrrect concept with $p(\theta^*|\text{prompt})=1$ and $p(\theta|\text{prompt})=0$ for all $\theta \in \Theta \setminus \theta*$. 
Even though we cannot locate the aligned concept, \citet{xie2021explanation} provides the theoretical guarantee on the effectiveness of the ICL in such cases, where the ICL performance improves along with the increasing number of in-context examples.

Inspired by the above Bayesian inference framework, more methods towards understanding skill recognition are proposed, e.g., the PAC-Bayesian framework~\cite{alquier2024user} and Hopfield Network~\cite{hopfield2007hopfield}. 
\citet{zhang2023and} analogizes ICL inference to a Bayesian model averaging algorithm. 
\citet{wies2023learnability} presents a PAC-based generalization framework exhibiting satisfying generalization bound on the ICL where a transformer trained on multi-task can match the ICL performance of a transformer trained solely on the downstream task.
\citet{zhao2023context} analogizes the latent concept location as memory retrieval with the Hopfield Network. 
More recently, a novel information-theoretic framework~\cite{jeon2024information} has been introduced, decomposing the ICL prediction error into three distinct terms: irreducible error, meta-learning error, and intra-task error.  
This decomposition helps aligning ICL with existing studies hypothesizing ICL as an instance of meta-learning.

Nonetheless, existing studies are based on either synthetic data or pure theoretical analysis. It could be a promising direction to investigate how LLMs retrieve concepts and how to interpret the retrieved concept through natural language.
\section{Skill Learning\label{sec:skill-learning}}

Through the skill learning ability, LLMs can inference a new data generation function which has not been seen during pre-training. 
The function learning framework\footnote{We refer to algorithm learning as function learning with an emphasis on the approximated functions by algorithms and, in this way, it is easier to analyze ICL.} is utilized to interpret the skill learning ability. 
Specifically, pre-training is considered as a process to learn a class of functions that can fit the pre-training corpora, and the ICL inference is to learn a new data generation function via fitting the ICL demonstrations. 

Discussions on the skill learning ability are organized as follows. 
In Section~\ref{subsec:stat-preliminary}, we first provide a clear description of the function learning framework and illustrate its benefits and drawbacks. 
In Section~\ref{subsec:stat-what}, we investigate: (1) whether LLMs can learn new functions in context, and (2) if so, whether the learned functions can effectively generalize to test samples. 

In Section~\ref{subsec:stat-how} illustrates ICL can implement different learning algorithms, e.g., gradient descent. 
More discussions on the robustness of ICL can be found in Appendix~\ref{app:stat-ood}.

\subsection{The Function Learning Framework  \label{subsec:stat-preliminary}} 
Previous research reformulates the pre-training objective of next-token prediction into an input-label mapping objective during the ICL inference stage.
One limitation of the function learning framework is that it has to pre-train the model from scratch as the pre-training objective is different from the next token prediction. 
Due to computational resource limitations, most works utilize transformers with less than 6 layers. 
These conclusions may not be generalizable to larger scale models.
\citet{garg2022can} has been the only work to utilize a relative larger-scale model, reaching a similar scale as GPT-2.

Denoting $\mathbf{x} \sim \mathcal{P}_{\mathcal{X}}, \mathbf{x}  \in \mathbb{R}^d$ where $\mathcal{P}_{\mathcal{X}}$ is a distribution, a function class $\mathcal{F}$ where for each $f \in \mathcal{F}, f: \mathbb{R}^d \to \mathbb{R}$.
Given a sequence ($\mathbf{x}_1, \cdots \mathbf{x}_i$) ($i > 1$) sampled from $\mathcal{P}_{\mathcal{X}}$ sequentially, and a sampled function $f \sim \mathcal{F}$, the learning objective aims to correctly predict $f(x_i)$ based on the sequence $(\mathbf{x}_1, f(\mathbf{x}_1), \cdots, \mathbf{x}_{i-1}, f(\mathbf{x}_{i-1}), \mathbf{x}_i)$ with both in-context examples and the query input $\mathbf{x}_i$.
\begin{equation} \label{eq:learn}
    \underset{\substack{\mathbf{x}_{1} \dots \mathbf{x}_{n} \sim \mathcal{P}_{\mathcal{X}} \\ f \sim \mathcal{F}}}{\mathbb{E}}\left[\sum_{i=2}^{n} \!\mathcal{L}\left(f\!\left(\mathbf{x}_{i}\right),T_{\omega}\left(\left[\mathbf{x}_{1},f\left(\mathbf{x}_{1}\right)\ldots \mathbf{x}_{i}\right]\right)\right)\right]
\end{equation}
Eq.~\eqref{eq:learn} describes the learning objective, where $\mathcal{L}$ is the loss function. 
$T_\mathbf{\omega}$ denotes the transformer model, $\mathbf{\omega}$ is the parameter of the transformer.

Notably, the model is pre-trained on the above ICL objective instead of the original next-token prediction objective. 
The function learning framework enables us to: 
(1) arbitrarily generate data with desired properties from the pre-defined function class $\mathcal{F}$; 
(2) clearly examine the function-approximation ability and the generalization of skill learning in ICL; and 
(3) utilize well-developed statistical learning theory to understand ICL.

\subsection{Function Approximation and Generalization of ICL\label{subsec:stat-what}}

In this subsection, we investigate the function approximation and generalization behavior of ICL. 
\textit{Function approximation} indicates to what extent transformers can approximate the ground-truth function underlying a given input, in the ICL inference stage.
\textit{Generalization}, on the other hand, measures the gap between the approximated function and the ground-truth data generation function. 
Notably, the function learning framework investigates ICL in the function space, rather than the token space.

To explore the function approximation ability, \citet{raventos2023pretraining} leverages different linear functions to generate pre-training data and in-context demonstrations. 
When pre-training on a small set of linear functions, ICL acts as a Bayesian optimal estimator, illustrating the skill recognition ability~\cite{raventos2023pretraining}. 
If enlarging the set of pre-training linear functions, ICL can act as an optimal least squares estimator with better function approximation, illustrating the skill learning ability~\cite{raventos2023pretraining}. 
\citet{wu2023many} provides a theoretical explanation to support the above empirical observations.

Beyond the linear function class, \citet{garg2022can} observes that the ICL is expressive enough to approximate more complicated functions, including sparse linear functions, two-layer neural networks, and decision trees.  The only requirement is that the same function class must be encountered during both pre-training and the ICL stage. 
\citet{bai2023transformers} and \citet{fu2023can} establish a statistical task complexity bound for pre-training, supporting the empirical observations mentioned above. The findings suggest that skill learning can be achieved with a dimension-independent number of linear regression pre-training tasks. 
However, two essential questions remain unsolved:
(1) Why do transformers suddenly obtain the skill learning ability with significant performance increase once the number of pre-training data generation functions reaches a certain threshold? 
(2) Why is the learned data generation function of ICL demonstrations from the same class as the pre-training data generation function?

The \textit{generalization} of ICL is validated by comparing the ground-truth data generation function of in-context demonstrations and the approximated one through ICL  inference. 
A more complicated experimental setting is considered where pre-training involves data generation functions from multiple function classes simultaneously, rather than being restricted to a single function class, as in the above function approximation experiments. 
Assuming pre-training data generation functions cover decision trees and linear functions, the ground-truth data generation function of ICL demonstrations is a linear function. The ICL generalization is strong if and only if the predicted function of ICL demonstrations is a linear one. 

\citet{bai2023transformers, ahuja2023context, vasudeva2024simplicity, tripuraneni2023can} indicate that transformers can achieve the Bayesian optimal selection, choosing the best-fitting function class with the minimum description length, from those function classes seen during the pre-training stage. 
Such Bayesian optimal selection helps a transformer pre-trained with multiple function classes reach comparable ICL performance as one pre-trained with only the ground-truth function class.
Notably, such Bayesian optimal on the synthetic dataset may not fully explain all the experimental observations. 
\citet{yadlowsky2023pretraining} generates each pre-training instance with functions from multiple function classes, e.g., $0.7 f_1(x) + 0.3 f_2(x)$ where $f_1$ and $f_2$ are from different function classes. 
The ICL can still achieve Bayesian optimal selection, holding the same conclusion.  
Notably, the above works focus on the scenario where the ground-truth data function is within pre-training function classes. 
Skill learning fails if the ground-truth data function is out of the pre-training function class~\cite{yadlowsky2023pretraining}; ICL degrades to skill recognition with Bayesian optimal estimator.

In summary, skill learning emerges if the number of pre-training data generation functions is sufficiently large.
ICL can learn a function that lies in the same function class of the pre-training data. 
Moreover, ICL would implement a Bayesian optimal selection to select the function best-fitting on ICL demonstrations, from pre-training function classes.

\subsection{The Internal Mechanisms of ICL \label{subsec:stat-how}}

In this subsection, we explore \textit{how ICL can learn an unseen function in context}. 
Notably, there are two common assumptions generally utilized in existing works: 
(1) The data generation functions for both pre-training data and in-context demonstrations are linear.
(2) The toy transformer model is linearized by removing feed-forward layers and the softmax activation function in the attention layer. 
This linearized simplification may generalize to the standard transformer, as \citet{ahn2023linear} illustrates that the training dynamic of the linearized version is similar to the standard transformer.

Previous works analogize ICL to meta-learning~\cite{finn2017model}. The pre-training stage corresponds to the outer-loop optimization, and the ICL inference stage is an instance of the inner-loop optimization, implementing fast adaptation on new novel tasks.
Rather than a real inner gradient update, ICL inference mimics gradient update via a forward process with in-context demonstrations~\cite{hubinger2019risks, von2023uncovering, zheng2024mesa}.

Based on the dual view that \textit{the backward process on a linear neural layer is equivalent to the forward process on a linear attention layer}, \citet{irie2022dual,dai2022can} proves the mathematical equivalence, illustrating the implicit gradient descent implementation with a linear attention.
However, such an analogy is only limited to mathematical equivalence. It remains unclear why ICL can learn a function since such an analogy overlooks many practical details, including the choice of the learning objective, pre-training weights, and the training data distribution~\cite{mahdavi2024revisiting}. 

To address the gap between theoretical models and real-world implementation, the following works consider the construction of pre-training weights.
\citet{von2023transformers} first demonstrate that ICL on the single-layer transformer can implement one-step gradient descent with a linear regression objective. 
\citet{bai2023transformers} further show that ICL inference can implement ridge regression, least square, lasso, and even gradient descent on a two-layer Neural Network. 
Nonetheless, those strong assumptions about the attention weights may be not practically reasonable.
For instance, \citet{von2023transformers} construct the key, query, value matrices $W_{K}, W_{Q}, W_{V}$ with 
$W_{K}=W_{Q}=\left(\begin{array}{cc}
I_{x} & 0 \\
0 & 0
\end{array}\right), W_{V}=\left(\begin{array}{cc}
0 & 0 \\
W_{0} & -I_{y}
\end{array}\right)$, where $I_{x}$ and $I_{y}$ are two different identity matrices and $W_{0}$ is the initialized parameters of the transformer model.
Nonetheless, it is unclear why a pre-trained transformer would have such type of weights, and it has been reported that this is not easily achieved in practice~\cite{shen2023pretrained}.

Instead of explicit attention weight construction, \citet{zhang2023trained, mahankali2023one, ahn2023transformers} analyze the \textit{converged weights} obtained after pre-training. 
\citet{von2023transformers} observes the ICL on the one-layer linear transformer can implement gradient descent or preconditioned gradient descent algorithm~\cite{ahn2023transformers} given a linear regression objective. Given a two-layer transformer, ICL can implement a gradient descent with adaptive step size and special sparsity regularization~\cite{ahn2023transformers}. 
Moreover, \citet{ahn2023transformers, von2023transformers} reveal that multiple-layered transformers can implement a GD++ algorithm. 
For larger-scale transformers, \citet{akyurek2022learning} empirically illustrates that, instead of performing GD, large-scale transformers show emergent ability directly approximating the closed-form solution of ridge-regression, while there is still a gap on why this ability emerges as the model-scale increases.

Beyond the linear activation for attention heads, recent researches take the softmax activation function into consideration. 
\citet{von2023transformers} demonstrates there exists a transformer that performs GD to solve more complicated nonlinear regression tasks.  
\citet{li2023shifting, ren2023context} identify the nonlinear regression task as the softmax regression and contrastive learning objective, respectively. 
\citet{cheng2023transformers} further takes non-linear data generation functions into consideration, elucidating a transformer can implement gradient descent and converge to the Bayes optimal predictor. 
\citet{wibisono2023role} theoretically finds that the softmax can help to find the correct data pair from the unstructured data which the input-output pair is permuted. 
\citet{guo2023transformers, zhang2024context} further studies a more challenging but practical setting of representation learning, in which predictions depend on inputs through the MLP. The theoretical evidence in \citet{guo2023transformers} indicates that the ICL inference can implement ridge regression in context with the input of neural representations, while \cite{collins2024context} argue theoretically and empirically that ICL inference with a single self-attention head behaves like a Nadaraya-Watson kernel regressor and training the attention weights entails learning the appropriate neighborhood size and subspace for this regressor based on the Lipschitzness of the target functions.

\textbf{Practical usage of mechanism analysis.} 
The above section has indicated that ICL implements a gradient descent  vector to achieve successful function learning. 
From a practical perspective, \citet{todd2023function, liu2023context} find the existence of compressed task vectors\footnote{Similar task vectors~\cite{hojel2024finding} can also be found in the computational vision domain.} in transformers with specific functionality. 
More recently, \citet{li2024context} attempts to connect the gradient vector with the compressed task vector, utilizing inner and momentum optimization towards a better task vector. Success of the new optimized task vector can be found on multiple tasks.

\section{Insights \& Future Directions \label{sec:insight}}

In this section, we delve into key insights from the data mechanism perspective of ICL and identify open questions that remain to be addressed in this evolving field.

\textbf{The uniformity of the two frameworks.} The core idea from the data-generative perspective is to (1) construct a data generation function hypothesis with one specific statistical framework and (2) analyze the data generation capability of the LLM with ICL instances with a focus on either skill learning/recognition mechanism. 
The existing pipelines on skill recognition and skill learning abilities are comprehensively discussed with the statistical frameworks of the Bayesian inference and function learning in Section~\ref{sec:skill-recognition} and~\ref{sec:skill-learning}, respectively. 
However, most existing analysis follows one-to-one correspondence which explains one ability with one specific statistical framework, serving as a solution for skill learning. 

Our new data generative perspective suggests the researcher find a suitable statistical framework as the starting point for analysis. We exhibit the potential that both frameworks can be easily utilized to understand the mechanism of both abilities. 
Such extension enables the future mechanism analysis to select the suitable analysis framework, by referring to their strengths and weaknesses. 
The function learning framework provides an elegant description of the data generation process with more comprehensive conclusions.
However, it is over-simplified with an unclear relevance to the real-world scenario. 
The Bayesian inference framework provides a more concrete and detailed description of the data generation process through an HMM model, e.g., the delimiter is taken into consideration, while the theoretical analysis on the role of delimiters is hard since it requires several assumptions over statistical modeling.

We provide a comprehensive discussion on extending one framework to the other statistical framework. 
The function learning framework can be easily extended to understand skill recognition by simply replacing the data generation function from a mixture of HMMs with linear functions. 
A comprehensive discussion on how to utilize the Bayesian inference framework to model the mechanism of skill learning in Appendix~\ref{sec:insight-framework}. 
We first show that the original function learning framework for the skill learning ability also implements an implicit Bayesian optimal selection in Appendix~\ref{subsec:tradeoff-algorithm}.  
We then extend the Bayesian inference framework to learn new in-context data generate functions in Appendix~\ref{subsec:tradeoff-bayesian}. The Bayesian inference framework can also serve as a solution for skill learning.

\textbf{The unique strengths and weaknesses of skill learning/recognition ability}
Considering the intricate interplay of both abilities on different tasks, we further illustrate the strengths and weaknesses inherent in each ability.
Skill learning ability can obtain new knowledge from the in-context data, and even over-ride the pre-training knowledge. 
It provides an easy way to update the knowledge on the specific application without requiring computationally heavy fine-tuning. 
Such ability has been successfully utilized in different LLM applications, e.g. model editing with ICL~\cite{zheng2023can}. 
Nonetheless, the skill learning ability may fail as it can be easily distracted by irrelevant context~\cite{shi2023large}. 
Skill recognition ability is insensitive to the new in-context pattern leading to the failure on the specification-heavy task~\cite{peng2023does} while it exhibits robustness to the incorrectness of label-demonstrations and other in-context noise~\cite{webson2021prompt}. 
Based on the above discussion, we suggest a careful evaluation of LLMs about each ability and select a desired one for the downstream task.

\textbf{Emergent Skill Composition Ability.} 
We majorly focus on the skill recognition/learning ability in our paper. 
More recently, new skill composition ability is found on larger model with specialized ICL prompts like Chain-of-Thought~(CoT)~\cite{wei2022chain}. 
The skill composition ability combines multiple data generation functions to create a more complicated data generation function. 
This ability, supported theoretically by \citet{arora2023theory}, shows that complex tasks can exhibit performance gains when decomposed skills improve linearly. More analyses on the effectiveness of skill composition ability can be found in Appendix~\ref{app:composition}. 

\textbf{Application of Skills.}
After the LLM obtained the skill learning and skill recognition abilities during pre-training, we then investigate how the model utilizes both abilities for achieving satisfactory downstream task performance during the ICL inference stage. 
Overall, the behavior of the LLM is more consistent with the skill recognition mechanism on difficult tasks while observations aligned with skill learning are more common to see on easy tasks.

Empirical analyses are conducted on the well-trained LLM, focusing on the ICL behavior on downstream tasks with various difficulties. 
Typically, we examine whether the model behavior aligns with the skill recognition ability or the skill learning one via the performance sensitivity on corrupting in-context data with incorrect input-label mapping. 
If the LLM takes advantage of the skill learning ability more, the LLM can learn the corrupted in-context mapping, leading to performance degradation compared with the origin setting. 
In contrast, if the LLM follows the skill recognition ability more, the LLM should be robust to the correctness of the input-label mapping, since the skill recognition ability only implements the pre-training data generation function with correct input-label mapping. 
\citet{min2022rethinking} first observes that the corrupted mapping does not necessarily lead to the overall performance degradation, indicating an overall skill recognition behavior. 
Instead of examining the overall performance across tasks, \citet{yoo2022ground} conducts a more careful evaluation of each task individually where the ICL shows different behaviors on tasks with different difficulties.  
The relatively easy tasks exhibit performance degradation on the wrong input-label mapping while the robust performance appears on those difficult tasks.  
Such observation indicates that the skill learning ability is more applicable to relatively easy tasks while the skill recognition ability dominates on the difficult ones.

\textbf{How the skill learning ability emerges during pre-training.} 
The emergence of the skill learning ability can be partially attributed to the skewed rank-frequency distribution of pre-training corpora. ~\cite{chan2022data}, and \cite{reddy2023mechanistic} highlight the role of the induction head~\cite{olsson2022context}, a particular attention head which explicitly searches for a prior occurrence of the current token in-context and copying the suffix as predictions. 
Moreover, the function class-based analysis~\cite{raventos2023pretraining} illustrates that the transition from skill recognition to skill learning only happens given diverse enough tasks in pre-training corpora.
It is interesting to explore how these factors collaboratively influence the emergence of skill learning.

\textbf{Why does ICL only learn the data generation function that appeared during pre-training?}
In Section~\ref{sec:skill-learning}, we provide a comprehensive discussion on what function can be learned in context. 
Observations indicate that ICL can only learn the function within the pre-training data generation function class. 
Nonetheless, the causality of the pre-training data generation function to ICL remains unclear. 
\citet{garg2022can} proposes the research question as: \textit{Can we train a model to in-context learn a certain function class} but overlooks the effect of the pre-training data generation function class. 
Once we have a certain clue about causality, we can leverage the skill-learning ability in a more controllable and safe manner.

Another line of research is to conduct analyses on more realistic scenarios. Recently, \citet{chen2024parallel} finds the parallel structures in pre-training data-pairs of phrases following similar templates in the same context window is the key to the emergence of the ICL capability. We conjecture that the underlying reason can be the formulation of the induction head with repeat patterns.

\textbf{Data generation functions aligned with real-world scenarios.} 
One major concern on the statistical framework is that the correspondence with real-world scenarios is unknown and overly simplified. 
Recently, \citet{akyurek2024context} proposes a new approach for generating data functions that are more aligned with real-world scenarios. 
The framework allows for more accurate simulations and testing of machine learning models by integrating domain-specific knowledge and constraints into the data generation process. 
This alignment enhances the applicability and reliability of existing conclusions to the real-world scenarios. 
We advocate for theoretical analyses focused on real-world data generation functions, moving beyond traditional statistical frameworks.
More empirical analysis on skill learning and skill recognition abilities are illustrated in Appendix~\ref{sec:empirical-tradeoff}.

\textbf{Extending existing findings to other capabilities of LLMs.}
More ICL capabilities are observed except for classification tasks, e.g., step-by-step reasoning ability~\cite{wei2022chain} for reasoning and self-correction~\cite{ganguli2023capacity}. 
A critical question is how we can extend the understanding frameworks introduced in this paper, particularly the data generation perspective, to more complicated LLMs' capabilities. 
Some pioneering research has been done; \citet{prystawski2023think} extends the Bayesian inference framework to understand the effectiveness of the CoT prompt. 
\citet{kadavath2022language} focuses on the self-evaluation prompt showing that LLMs can accurately examine the correctness of their statements.
We believe the introduced data generation perspective and two main understanding frameworks on ICL serve as the milestone to explore more intrinsic capabilities of LLMs. 

\section{Conclusion}

In this study, we introduce a novel data generation perspective to understand the underlying mechanism driving the current success of ICL. 
We primarily focus on understanding the LLM's ability of skill learning and skill recognition, and investigate whether ICL inference is capable of learning new data generation functions in context. 
Our work makes a step forward to enhancing our understanding of underlying mechanisms.

\section{Limitations}
\label{sec: bi}

In this paper, we provide a mechanism understanding of the ICL from a data generation perspective, 
We systematically consider the limitations from various perspectives such as fairness, security, harm to people, and so on, and we do not find any apparent social risk related to our work. 
However, there is a notable technical limitation in our study. 
The current statistical frameworks with controlled experimental settings may not fully capture complexities present in real-world scenarios. 
This gap between the theoretical framework and practical applications suggests that further research is needed to adapt and refine the mechanism analysis to align with real-world application. 

\vspace{-0.5em}
\section*{Acknowledgement} 
\vskip -0.5em

We extend our gratitude to Dr. Liam Collins for his insightful feedback on this paper.
Haitao Mao and Jiliang Tang are supported by the National Science Foundation under grant numbers CNS2321416, IIS2212032, IIS2212144, IOS2107215, DUE2234015, CNS2246050, DRL2405483 and IOS2035472, the Army Research Office under grant number W911NF-21-1-0198, Amazon Faculty Award, JP Morgan Faculty Award, Meta, Microsoft and SNAP.

\bibliography{main}
\newpage
\appendix
\section{Insights on the Bayesian Inference and the Function Learning Framework \label{sec:insight-framework}}

\subsection{Bayesian Selection in the Function Learning Framework \label{subsec:tradeoff-algorithm}}
The Bayesian perspective can be found in the function learning framework originally utilized for the skill learning mechanism. 
Typically, we illustrate the underlying Bayesian selection in the function learning framework, indicating the intrinsic connection between the two statistical frameworks. 
According to~\citet{ahuja2023context}, the transformers pre-trained on the data generated from diverse function classes exhibit improved function-fitting ability across all the pre-training function classes. 
To identify the best-fit solution among the whole function class, the function selection process implements a Bayesian optimal selection. More details can be found in Section~\ref{subsec:stat-what}. 
Notably, instead of the original Bayesian inference framework only selecting pre-training data generation functions, the function selection scope is enlarged, including all the unseen functions from the same function class with the pre-training functions.

\subsection{Extending the Bayesian Inference Framework for Skill Learning\label{subsec:tradeoff-bayesian}}
We then illustrate the possibility of extending the Bayesian inference framework to understand the skill learning mechanism to capture new data generation functions from the in-context data via relaxing the particular assumption. 
One important assumption in the Bayesian inference framework~\cite{xie2021explanation} is that all ICL demonstrations should be generated with the same latent concept. 
Nonetheless, this strong assumption may not be held in practice. For instance, one demonstration sample discusses the topic of sociology but another one is relevant to cardiology, the data generation function for these two domains should be rather different. 
Inspired by the high compositionality nature of language data, \citet{hahn2023theory} came up with an information-theoretic bound showing that ICL performance can be improved given more unique compositional structures in pre-training data, therefore skill learning ability can appear by combining compositionality structures, in pre-training data, to infer the data generation function of ICL demonstrations.

Empirical evidence shows that, given an input-label pair of two semantically unrelated concepts, e.g., mapping sports to animals,~\citet{rong2021extrapolating, wei2023larger} still observe a satisfactory performance with the increasing model scale, indicating that the LLM can retrieve multiple concepts and combine them as a new data generation process. 
\citet{feng2023language} interpret the combination with a binding mechanism with an internal function vector to recognize the input feature and bind it to the corresponding label.

\citet{swaminathan2023schema} proposes another way to extend the existing Bayesian framework for skill learning via replacing the original HMM model into the clone-structured causal graph~(CSCG)~\cite{george2021clone, dedieu2019learning}. 
The major difference is that the CSCG considers a learnable emission matrix, which determines the probability of observing a particular output given each hidden state in the model. 
A relevant transition matrix as the concept is retrieved, similar to the Bayesian inference \cite{xie2021explanation}. 
The hidden states for each token can then be obtained given the particular relevant template. 
The LLM then learns the suitable emission matrix, providing the best-fit mapping from the hidden states to the observed token.

\section{Empirical Investigation On Skill Recognition and Skill Learning\label{sec:empirical-tradeoff}}

In this section, we exhibit more empirical analyses revolving around skill recognition and skill learning abilities. 
In contrast to the mechanism analysis that focuses on whether the ICL can learn new in-context data generation functions or not, empirical evidence in this section indicates that it is highly likely that LLMs exhibit both skill recognition and skill learning abilities of various levels, instead of an all-or-nothing conclusion. 
We first discuss how the LLM jointly obtains both abilities during the pre-training stage in Section~\ref{subsec:tradeoff-origin}. 
Specifically, the origin of both abilities is determined by the pre-training data distribution~\cite{chan2022data} and the model scale~\cite{wei2023larger, pan2023context}. 
Typically, the LLM exhibits varying degrees of usage on those two abilities according to tasks with different difficulties.

\subsection{Origin of Skills\label{subsec:tradeoff-origin}}

In this subsection, we carefully examine how well the LLM obtains the skill learning and the skill recognition abilities during the pre-training stage, with a focus on the impact of the pre-training data distribution and model scale. 
Roughly speaking, the skill recognition ability is easy to achieve while the skill learning ability develops much slower and only emerges when the model scale is sufficiently large. 

Analyses are first conducted focusing on how those abilities are developed along the pre-training procedure. 
\cite{bietti2023birth} observe that the skill recognition ability is obtained early in the pre-training procedure, while the skill learning ability is developed much later.
However, \citet{singh2023transient} shows that the obtained skill learning ability gradually vanishes after over-training and is replaced by the skill recognition ability.
Such observation indicates that skill learning is a transient ability that may disappear when the model is over-trained rather than a persistent one which can be kept once obtained.
The reason can be attributed to the pre-training data distribution~\cite{chan2022data} where the task learning ability degrades if the pre-training data follows a uniform, i.i.d distribution.
Nonetheless, such degradation may not happen when the pre-training data follows a properly skewed Zipfian distribution. 
\citet{chan2022data} further emphasizes that the skill learning ability emerges when the pre-training data meets the following properties: 
(1) Skewed rank-frequency distributions: Dynamic contextual meaning does not uniform across data, instead, only a few meanings dominate with the long tail of other infrequent meanings. 
(2) Burstiness: Dynamic contextual meaning is not uniform across time, but appears in clusters. 
The reason why ICL ability can be obtained on such data distribution remains unclear. A potential explanation could be that the pre-training weight can only obtain the head meaning frequently appears while the long tail knowledge can only be obtained via ICL.

Analyses are then conducted with a focus on the impact of the model scale.
\citet{pan2023context} illustrates that the skill recognition ability can be found across LLMs with different scales.
In contrast, LLMs obtain better skill learning ability along with an increasingly larger scale. 
Similar observations can be found in~\cite{wei2023larger} that the LLM can learn the flipped input-label mapping and override pre-training knowledge when the model scale is sufficiently large. 
\cite{fu2023does} provides the potential explanation where the good skill recognition ability serves as a necessity for developing the skill learning ability.


\section{Skill Composition \label{app:composition}}
We primarily focus on the skill learning ability where the ICL can learn a new data generation function, and skill recognition ability where the ICL utilizes the data generation function from pre-training data. 
Instead of focusing on the single data generation function, combining multiple data generation functions together can lead to a complicated data generation function. 
We named such capability as skill composition capability, helping the LLM to achieve a complicated task by combining a sequence of simple and basic steps. 
\citet{arora2023theory} theoretically indicates the effectiveness of skill composition where the complicated task can exhibit emergent performance gain when all the decomposed basic skills improve linearly.

The discussions on skill composition are organized as follows. 
In Section~\ref{subapp:com-effective}, we investigate the effectiveness of skill composition ability. 
In Section~\ref{subapp:com-when}, we analyze when the skill composition capability can work. 
In Section~\ref{subapp:com-discuss}, we further illustrate more discussion and real-world applications on the skill composition ability. 
Notably, the skill composition ability is complicated without a general data generation function framework so far. 
The skill-composition ability often requires to be elicited by specific-designed ICL prompts, e.g., Chain-of-Thought prompting~(CoT)~\cite{wei2022chain}, Tree-of-thought~\cite{yao2023tree}, and Graph-of-Thought~\cite{besta2023graph}, which generates multiple intermediate steps before the final answer. 
Most following literature conducts analysis on the CoT prompt.

\subsection{Effectiveness of Skill Composition\label{subapp:com-effective}}

In this section, we investigate the effectiveness of skill composition ability. 
\citet{feng2023towards} indicates that if the skill decomposition is applied, the LLM can be more expressive to describe more complicated problems, e.g., mathematical and decision-making problems. 
\citet{li2023dissecting, yang2023explaining} further demonstrate the data efficiency where the skill composition facilitates can learn complicated functions with a reduced sample complexity.  
\citet{prystawski2023think} attributes the above expressiveness and efficiency with the local structures in the training data generation function. 
Such locality enables to accurate inference on each intermediate step supported by the  similar pre-training data generation function. 
In contrast, direct inference as a whole instead of each local steps are likely to fail requiring since such complicated data generation function does not appear during the pre-training stage. 
In summary, the skill composition ability of LLMs enhances their expressiveness and data efficiency for modeling complicated data generation function, building on the basis of locality data generation function from the pre-training data.

\subsection{When Skill Composition Works \label{subapp:com-when}}

We demonstrate the effectiveness of the composition in Section~\ref{subapp:com-effective}, however, it remains unknown whether the decomposed intermediate steps are well-organized aligning with human cognition. 
To examine the correctness of the LLM decomposition, the literature focuses on formal deductive reasoning tasks like math reasoning~\cite{ahn2024large}. 
It enables to conducting systematic and controllable analysis on each reasoning step with the unique correct answer.

LLMs are able to conduct correct decomposition on particular tasks, aligning with the ideal human reasoning process. 
\citet{zhou2023algorithms} finds a theoretical criterion to identify when the LLM can implement the ideal decomposition. 
Typically, when the task can be described by a short RASP program~\cite{weiss2021thinking}, a programming language designed for the computational model of a Transformer, the LLM can achieve the correct decomposition. 
Similarly, \citet{yao2021self} demonstrates that the transformer can process correct decomposition on particular formal languages with hierarchical structure, e.g., $\text{Dyck}_k$~\cite{chomsky1959algebraic}. 
With a suitable decomposition, LLMs can easily solve arbitrary complicated problems~\cite{jelassi2023length, li2023representations}.

Beyond those identified tasks, it remains many tasks where LLMs cannot conduct an ideal decomposition. 
The key underlying reason~\cite{mccoy2023embers} is the gap between human cognition and the next-token prediction pre-training task, requiring to tackle problems sequentially greedily. 
Instead of a proper decomposition, a greedy shortcut can be obtained from standard training, which skips the particular step instead of a formal decomposition. 
Theoretical evidence on the existence of shortcuts can be found in~\cite{liu2022transformers} on the semi-automaton reasoning task. 
\citet{saparov2022language} indicates that the shortcut can easily select the wrong step, leading to an incomplete planning and subsequently an incorrect answer, leading to failure on complicated tasks~\cite{dziri2023faith}. 
Such inherent failure is unavoidable as the transformer always finds a shortcut solution~\cite{liu2022transformers} while impossibile to find the exact implementation of the semi-automaton reasoning requiring recurrent models of computation with shallow and non-recurrent architecture.
On the contrary, the shortcut also shows its benefits, converting the original complicated reasoning problem with multiple hops into a simpler one with less hops~\cite{wu2023reasoning, saparov2022language}, alleviating the performance degradation along with the increased hop. 

In summary, the shortcut solution of LLMs can be a double-side sword to solve a compositional problem. 
Nonetheless, it remains no existing study on how the LLM acquires the decomposition capability from pre-training data.  
Notably, we focus on whether the LLM composition aligns with the human decomposition while the manually-conducted deduction rules may not be optimal. 
The optimal decomposition remains unknown.

\subsection{More Discussions\label{subapp:com-discuss}}

Despite the above comprehensive understanding, there are more empirical studies on the skill composition ability from various perspectives as follows.
\citet{madaan2022text} divides the CoT prompt into three key components: symbols, patterns, and text with distinct roles as follows: 
(1) The exact type of symbols does not matter. 
(2) The patterns are the template serving as a trigger helping to locate the correct concept
(3) Text contains commonsense knowledge and meaning, leading to the ultimate success. 
Similarly, \citet{wang2022towards} divides the CoT prompt into two key components: bridging objects (the key and necessary objects) and language templates. Interestingly, neither of them matters. In contrast, the relevance to the query and correct reasoning ordering matters.

More recently, \citet{xu2024large} challenges the skill compositional capability of LLMs, pointing out the failure on the sequential reasoning tasks. On the contrary, LLMs can perform well on simple composite tasks that can be easily separated into sub-tasks based on the inputs solely. The skill composition ability remains mysterious, requiring further analyses.

\section{Transformer architecture simplification \label{app:simplification}}

To facilitate analysis, many studies introduce necessary simplifications to the standard Transformer architecture. While all empirical analyses use the standard Transformer setup, theoretical analyses adopt a modified version without layer normalization. More detailed theoretical simplification can be found as follows.

\begin{itemize}
    \item Do not consider model architecture~\cite{xie2021explanation, hahn2023theory}  
    \item A single-layer linear attention~\cite{von2023transformers, ahn2023transformers, mahankali2023one} 
    \item A single-layer relu attention~\cite{fu2023can}
    \item A single-layer softmax attention~\cite{zhao2023context, zhang2023and, ren2023context, li2023shifting} 
    \item An L-layer linear attention~\cite{ahn2023transformers}  
    \item A single-layer linear attention with FFN~\cite{von2023transformers}  
    \item A full transformer~\cite{akyurek2022learning, cheng2023transformers, bai2023transformers, guo2023transformers} 
\end{itemize}

\section{Discussions\label{app:discussion}}

\subsection{The Emergence Phenomenon On the ICL Generalization}

\citet{chan2022transformers} proposes an interesting perspective to characterize how the ICL generalizes to the test data based on the in-context samples. 
Observations exhibit that the larger LLMs can achieve rule-based generalization similarly with the SVM. 
The rule-based generalization makes decisions using a minimal set that is central to the category definition, disregarding less essential data, 
Nonetheless, induction heads mechanism with prefix match and copy are more aligned with examplar-based generalization like KNN. 
The reason why LLM can achieve rule-based generalization still remains unclear.

\subsection{Advantages And Disadvantages of Skill Learning And Skill Recognition}

Skill learning mechanism can obtain new knowledge from the in-context pattern, and even over-ride the pre-training knowledge. 
It provides an easy way to update the knowledge on the specific application without requiring computational-heavy fine-tuning. 
Such ability has been successfully utilized in different LLM applications, e.g. model editing with ICL~\cite{zheng2023can}. 
Nonetheless, the skill learning mechanism may fail as it can be easily distracted by irrelevant context~\cite{shi2023large}. 
The failure reason found in~\cite{tang2023large1} is that the input-label mapping is more to be the shortcut as the model scale increases.
Skill recognition mechanism is insensitive to the new in-context pattern leading to the failure on the specification-heavy task~\cite{peng2023does} while it exhibits robustness to the incorrectness of label-demonstrations and other in-context noise~\cite{webson2021prompt}. 
For instance, the skill recognition mechanism can perform well in a noisy setting as it can only locate the origin ability developed during the training procedure. The LLM cannot learn the new in-context information with noisy labels. Instead, it only helps to locate the most similar concept seen during the pre-training stage. Despite the labels being noisy, ICL may still be able to locate the correct concept with the input text information. Empirical evidences~\cite{min2022rethinking} indicates that even random permute the model label can lead to a satisfying performance.

\subsection{Abstraction Ability of LLMs} 

Despite the success of LLM based in the natural language, \cite{webb2023emergent, mirchandani2023large, huang2023lexinvariant, chen2023improving} indicate the effectiveness on abstract symbol without knowing semantic meanings of any individual symbol. 
\citet{webb2023emergent} exhibits the emergence ability of LLM for abstract pattern induction while \cite{mirchandani2023large} suggest that LLM is a general pattern machine extrapolating sequences of numbers that represent states over time to complete simple motions.  
\citet{huang2023lexinvariant} achieves comparable performance using random Gaussian vectors instead of the original token embedding when context is sufficient. 
\citet{chen2023improving} indicates such abstraction with randomizing embeddings can help LLM learn multiple languages.

\subsection{Discussion On the Self-correction}
The self-correction~\cite{pan2023automatically, kim2023language, gou2023critic, welleck2022generating} is an advanced ICL technique iteratively revise the outputs of LLM utilizing feedbacks, aiming to mitigate undesired and inconsistent behaviors, e.g., lexically constrained generation and toxic reduction. 
Despite its effectiveness, the underlying mechanism remains an open question. The initial observations can be found as follows.  
\citet{kadavath2022language} illustrates positive evidence where LLM can accurately examine the correctness of their statements, serving as the necessary condition for self-correction.
Nonetheless, \citet{huang2023large} observes that self-correction cannot improve the performance since the added feedback may bias the model away from producing an optimal response to the initial prompt.
\citet{hong2023closer} provides more detailed evaluation setting and identifies that (1) LLMs perform much worse at identifying fallacies related to logical structure than those related to content. (2) LLMs cannot classify different types of fallacies. 
Despite the above phenomenons, there is still no understanding of the underlying mechanism of self-correction so far.

\subsection{How The Data-generating Functions Are Different Than Arbitrary Functions}

We first emphasize the importance of the data generation function. 
The strong generative capability is an essential ability for LLMs. Most successful applications and usage of the LLM revolve around the generative capability. Therefore, the data generation perspective is essential to understand the LLM. 

The data-generating function is generally utilized to understand the data-generation capability of LLMs. 
It can be defined as 'the underlying hypothesis on textual data generation'. 
Technically, the data-generation function can be any function that can model the probability over a potential token given a sequence of tokens, after being trained with text data. 
The main difference between the data-generation function and arbitrary function is whether the function can be used to generate reasonable natural language sequences. Understanding the data-generation process is a core problem in natural language processing, particularly for natural language generation tasks.

More concretely, N-gram, HMM, and Recurrent Neural Networks are three straightforward data-generation functions but they cannot model long contexts, and the first two are non-parameterized data-generation functions. On the other hand, we can have a linguistic-driven data generation function, e.g. probabilistic context-free grammar~\cite{hahn2023theory}, to introduce some priors of syntax. Since the complicated and hierarchical nature of human languages, LLMs are great in terms of incorporating contextual information through a powerful function approximation ability. Honestly speaking, we can claim that the impressive results of LLMs depend on the ability to approximate the unknown data-generation function underlying the pre-training corpora.

Notably, the statistical framework, which utilized the input-label mapping as the data generate function is a simplified setting. 
Such a simplified setting enables to conduct of more theoretical analysis. Therefore, we can qualitatively analyze the expressiveness,  generalization, and internal mechanisms of the ICL. For instance, with the function abstraction, we can analyze the generalization within the same function class and between different function classes.
However, how to take advantage of it in a real-world scenario remains unclear.

\subsection{Whether Different Demonstrations Represent Different Data Generation Functions}

Whether different demos represent different data generation functions depends on the hypothesis of the data generative function. It is possible for different demonstrations to share the same data generation function. On the contrary, it is also possible for different orders of the demonstrations to correspond to different data generation functions.

\subsection{Whether there is the connection between skill learning/recognition and model under/overfitting?}

The ICL procedure does not have any backward learning process, i.e. gradient descent, generally utilized in deep learning. Therefore, the ICL procedure is not explicitly related to the model under/overfitting without an explicit fitting procedure. 

Both skill learning and skill recognition can achieve a certain generalization, without explicit under-fitting or over-fitting. 
The skill recognition is not directly memorization. Given the train data $(\mathbf{x},\mathbf{y})$ generated from the function $\mathbf{y}=\mathbf{k}\mathbf{x}$, the pre-training data can be within the input interval $\mathbf{x} \in [ 0,1 ]$, while the ICL test data can be within the input interval $\mathbf{x} \in [1,2]$. In such a case, the ICL can still achieve satisfying performance, indicating the generalization ability. It indicates the ICL with skill recognition can achieve generalization when test data are within the same function. 
A more comprehensive discussion when meeting out-of-distribution scenarios can be found in Appendix~\ref{app:stat-ood}. 

The difference between skill learning and recognition is the different extent of the generalization. 
The skill recognition generalizes through seeking an existing function within the same function class but skill learning can come up with a new function within this function class.

\subsection{The real-world correspondence of data generation functions}

Our paper focuses on whether the ICL can learn a new data generation function in context.
From a practice scenario, the new data generation function can be defined as the n-gram does not appear in the training stage. Such compositional generalization is a key concept in the NLP domain. For instance, such out-of-distribution can happen when LLMs read the news. The skill learning mechanism can learn the new n-gram and knowledge in context, while skill recognition tries to map the pre-training knowledge with the news. 
\section{The Robustness of ICL On the Statistical Framework\label{app:stat-ood}}

We primarily analyze the skill-learning mechanism when (1) data generation functions during the pre-training and ICL inference stages are from the same function class, and (2) input features are sampled from the same distribution in Section~\ref{sec:skill-learning}. 
In this section, we provide a further discussion of how the skill-learning mechanism works when distribution shifts happen, indicating the robustness of the ICL. 
The robustness of the ICL is evaluated in different out-of-distribution scenarios, which can be roughly divided into the following categories: 
(1) Task shift, where the pre-training and in-context labels are generated from different function classes, is discussed in Appendix~\ref{subapp:task-shift}. 
(2) Corvariate shift, where the pre-training and in-context inputs are sampled from different distributions, is discussed in Appendix~\ref{subapp:corvariate-shift}.   
(3) Query shift, where the in-context training inputs and the query sample input are sampled from different distributions, is discussed in Appendix~\ref{subapp:query-shift}. 
Notably, all the above out-of-distribution scenarios are conducted on the statistical framework while it remains an unclear correspondence to the real-world LLM system pre-training on the massive corpus. 
More recently, \citet{vladymyrov2024linear} focuses on the corrupted training data scenario with noises on different extend. Both empirical and theoretical results indicate the robustness of transformers in such scenario.

\subsection{Preliminary}

To formally describe different out-of-distribution scenarios, we first provide a rigorous description of the pre-training and prompt data from a distribution perspective. 
The pre-training data is defined as $(\mathbf{x}_1, \mathbf{h}(\mathbf{x}_1), \cdots , \mathbf{x}_N, \mathbf{h}(\mathbf{x}_N), \mathbf{x}_{\text{query}})$ where $\mathbf{x}_i\sim \mathcal{D}_\mathbf{x}^{\text{train}}$, $\mathbf{x}_{\text{query}} \sim \mathcal{D}_\mathbf{x}^{\text{train}}$ and $\mathbf{h} \sim \mathcal{D}_\mathcal{H}^{\text{train}}$. 
The test prompt is defined similarly but drawing from a different distribution where $\mathbf{x}_i \sim \mathcal{D}_{\mathbf{x}}^{\text{test}}$ and $\mathbf{x}_{\text{query}} \sim \mathcal{D}_{\mathbf{x}}^{\text{test}}$. 
We then describe different out-of-distribution scenarios and how the LLM behaves on them differently in the following sections. 

\subsection{Task Shift\label{subapp:task-shift}} 

Task shift~\cite{zhang2023trained} is a concept shift which be formally defined as $\mathcal{D}^{\text{train}}_{\mathcal{H}} \ne \mathcal{D}^{\text{test}}_{\mathcal{H}}$.
It describes that the pre-training and in-context labels are generated from different function groups. 
Existing literature demonstrates two different task shifts, i.e., noise shift~\cite{zhang2023trained}, and regression vector shift~\cite{raventos2023pretraining}. 

Noise shift~\cite{zhang2023trained} corresponds to the scenario where the shift is induced by the random Gaussian noise. 
Typically, the pre-training data generation function is $\mathbf{y}=\left \langle \mathbf{w}, \mathbf{x} \right \rangle$ where in-context data generation function is from noisy linear function $\mathbf{y}_i=\left \langle \mathbf{w}, \mathbf{x}  \right \rangle + \mathbf{\epsilon}$. 
\citet{zhang2023trained} observes satisfying performance under such shift, indicating the robustness under such Gaussian noise. 

Regression vector shift~\cite{raventos2023pretraining} corresponds to the scenario where pre-training data generation functions are a limited group $\mathcal{F}_{\text{train}}$ of linear functions $\mathbf{f_i}: \mathbf{y}=\left \langle \mathbf{w}_i, \mathbf{x}  \right \rangle + \mathbf{b}_i$, where $\mathbf{f_i}\in \mathcal{F}_{\text{train}}$
The in-context data generation function is from all the possible linear functions covering the entire function space $\mathbf{f}_i \in \mathcal{F}_{\text{context}}$, where $\mathcal{F}_{\text{train}} \subseteq \mathcal{F}_{\text{context}}$. 
The task shift appears on the unseen data generation function during training.  
\citet{raventos2023pretraining} observes that ICL exhibits the generalization gap with insufficient pre-training data.
The emergence happens when the number of pre-training functions increases with satisfying out-of-distribution performance.

\subsection{Covariate Shift\label{subapp:corvariate-shift}}

Covariate shift~\cite{zhang2023trained} can be formally defined as $\mathcal{D}^{\text{train}}_{\mathbf{x}} \ne \mathcal{D}^{\text{test}}_{\mathbf{x}}$. 
It describes that the pre-training inputs and the in-context inputs are sampled from different distributions. Existing literature demonstrates different covariate shifts including low-dimensional subspace shift, skewed covariance shift, mean shift, and random covariate shift. 

Low-dimensional subspace shift~\cite{garg2022can} samples prompt input feature from random 10-dimensional subspace from the pre-training input feature. 
\citet{garg2022can} empirically observes the robustness over such covariate shift. 

Skewed covariance shift~\cite{garg2022can} samples in-context features from $\mathcal{N}(\mathbf{0}, \mathbf{\Sigma})$ where $\mathbf{\Sigma}$ is a skewed covariance matrix with eigen-basis chosen uniformly at random and $i^{\text{th}}$ eigenvalue proportional to $1/i^2$. 
Empirically observations~\cite{garg2022can} indicate the performance degradation when the input feature dimension is larger than 10. 

Mean shift~\cite{ahuja2023closer} samples train and test inputs from $\mathcal{N}(\mathbf{\mu}_{\text{train}}, \mathbf{\Sigma})$ and $\mathcal{N}(\mathbf{\mu}_{\text{test}}, \mathbf{\Sigma})$ where $\mathcal{N}(\mathbf{\mu}_{\text{train}} \ne \mathcal{N}(\mathbf{\mu}_{\text{test}})$. 
Despite performance degradation to a certain extend, the transformer backbone shows better generalization than the MLP backbone with both empirical observations and theoretical evidence. 

Random covariate shift~\cite{zhang2023trained} corresponds to that pre-training training prompts and in-context prompts are sampled from distributions with different covariates. 
The ICL performance degradation~\cite{von2023transformers, zhang2023and} drops to 0 quickly with theoretical explanation~\cite{zhang2023and}. 
The larger transformer with non-linearity serves as the solution to random covariate shift, while the reason underlying the emergent ability remains unclear.

\subsection{Query Shift\label{subapp:query-shift}}

Query shift~\cite{zhang2023trained} is the covariate shift, which can be formally defined as $\mathcal{D}^{\text{test}}_{\text{query}} \ne \mathcal{D}^{\text{test}}_{\mathbf{x}}$. 
It describes the distribution shift within the in-context training samples and test samples are sampled from different distributions. 
Different from the task shift focusing on the distribution shift between pre-training data and prompt data, query shifts describe the distribution shift within the prompt data, where the training prompt data distribution is different from the prompt query distribution. 
Existing literature demonstrates two different query shifts as follows.

The orthants shift changes the positive or negative signs to each coordinate of in-context features, ensuring both prompt data and prompt query fall within the same orthant, distinct from the query input's orthant. 
\citet{garg2022can} observes the robustness to this shift when differences between orthants are not large.

The orthogonal shift maps the the prompt query to the orthogonal space of prompt data, which is an extreme case of the formal one. 
\citet{garg2022can} shows empirical evidence where the prediction will be zero and the error will be significantly large.
\citet{zhang2023and} further theoretically underpins the underlying reason while no solution is found currently.

\end{document}